# Vessel segmentation for $\chi$-separation

Taechang Kim[1], Sooyeon Ji[1,2], Kyeongseon Min[1], Minjun Kim[1], Jonghyo Youn[1], Chungseok Oh[1], Jiye Kim[1], Jongho Lee[1,*]

**Author affiliations:**

[1]Laboratory for Imaging Science and Technology, Department of Electrical and Computer Engineering, Seoul National University, Seoul, Korea, Republic of

[2]Division of Computer Engineering, Hankuk University of Foreign Studies, Yongin, Korea, Republic of

**Correspondence author:**

Jongho Lee, Ph.D

Department of Electrical and Computer Engineering, Seoul National University

Building 301, Room 1008, 1 Gwanak-ro, Gwanak-gu, Seoul, Korea

Tel: 82-2-880-7310

E-mail: jonghoyi@snu.ac.kr

**Word count**: 4305

# Abstract

**Purpose:** $\chi$-separation is an advanced quantitative susceptibility mapping (QSM) method that is designed to generate paramagnetic ($\chi_{para}$) and diamagnetic ($|\chi_{dia}|$) susceptibility maps, reflecting the distribution of iron and myelin in the brain. However, vessels have shown artifacts, interfering with the accurate quantification of iron and myelin in applications. To address this challenge, a new vessel segmentation method for $\chi$-separation is developed.

**Methods:** The method comprises three steps: 1) Seed generation from $R_2^*$ and the product of $\chi_{para}$ and $|\chi_{dia}|$ maps; 2) Region growing, guided by vessel geometry, creating a vessel mask; 3) Refinement of the vessel mask by excluding non-vessel structures. The performance of the method was compared to conventional vessel segmentation methods both qualitatively and quantitatively. To demonstrate the utility of the method, it was tested in two applications: quantitative evaluation of a neural network-based $\chi$-separation reconstruction method ($\chi$-sepnet-$R_2^*$) and population-averaged region of interest (ROI) analysis.

**Results:** The proposed method demonstrates superior performance to the conventional vessel segmentation methods, effectively excluding the non-vessel structures, achieving the highest Dice score coefficient. For the applications, applying vessel masks report notable improvements for the quantitative evaluation of $\chi$-sepnet-$R_2^*$ and statistically significant differences in population-averaged ROI analysis. These applications suggest excluding vessels when analyzing the $\chi$-separation maps provide more accurate evaluations.

**Conclusion:** The proposed method has the potential to facilitate various applications, offering reliable analysis through the generation of a high-quality vessel mask.

# 1. Introduction

Iron and myelin have critical roles in normal brain functions,[1,2] and alterations in their levels are often associated with neurodegenerative diseases such as Parkinson's disease (PD), Alzheimer's disease (AD), and multiple sclerosis (MS).[3–7] These changes highlight the potential of iron and myelin as biomarkers for the diseases, suggesting the need for imaging techniques that can effectively delineate the spatial distribution of iron and myelin.

$\chi$-separation (chi-separation or x-separation) is an advanced quantitative susceptibility mapping (QSM) method that generates paramagnetic and diamagnetic susceptibility maps, potentially reflecting the distribution of iron and myelin in the brain.[8,9] This method combines local field information with reversible transverse relaxation rates ($R_2'$) to separate the paramagnetic and diamagnetic susceptibility sources within a voxel. The method has been applied to various studies in evaluating neurological disorders.[10–16] Additionally, similar susceptibility source separation techniques have been developed, expanding the scope of potential applications.[17–20]

In $\chi$-separation, vessels have shown to create erroneous artifacts, hampering its applications.[9,21] Large susceptibility differences around veins can induce mesoscopic field inhomogeneities,[22,23] creating non-local $R_2'$ (or $R_2^*$) effects. These effects contradict with the assumption that $R_2'$ (or $R_2^*$) effects are fully localized in each voxel,[9] resulting in artifacts near the veins. Furthermore, flow inside the vessels causes spatial displacement,[24,25] leading to inconsistent signal decay across echo times within and near the vessels. This inconsistency results in inaccurate $R_2^*$ values, propagating errors in $\chi$-separation maps.[21] These vessel artifacts can interfere with the accurate quantification of iron and myelin in some applications.[26,27] For reliable analysis, the exclusion of vessels via vessel segmentation is an option.

Until now, several vessel segmentation methods have been developed and applied to susceptibility imaging such as susceptibility weighted imaging (SWI) and QSM for disease assessment or oxygen extraction fraction (OEF).[28–37] Among them, Hessian-based vessel enhancement filters, such as Frangi filter,[38] stand out due to their effectiveness in highlighting tubular structures. The eigenvalues of the second order derivatives matrix (Hessian matrix) of the image intensity have been used to calculate the "vesselness" which indicates the likelihood of each voxel belonging to a vessel.[38] Based on a similar idea, a few important improvements have been made.[39–42] However, applying only Hessian-based vessel enhancement filters to susceptibility images poses challenges in differentiating vessels from regions with high susceptibility concentration such as deep brain nuclei. To address this challenge, new methods have been proposed to apply the Hessian-based vessel enhancement filter and then apply additional steps to reduce false positive voxels.[43,44]

This study proposes a new vessel segmentation method for $\chi$-separation. The method incorporates the physics of $\chi$-separation and the geometry characteristics of vessels such that it creates a high-quality vessel mask without non-vessel structures. The method is expected to facilitate various applications, providing reliable analysis.

## 2. Methods

**2.1** Proposed vessel segmentation method

The overview of the proposed method is illustrated in Fig. 1. The method comprised three steps. In the first step, seeds for vessels were obtained. Then, region growing, guided by vessel geometry characteristics, was applied to generate a vessel mask. In the final step, this mask was refined by removing non-vessel structures. This method utilized four inputs: $R_2^*$, paramagnetic susceptibility ($\chi_{para}$) and diamagnetic susceptibility ($|\chi_{dia}|$) maps, and a brain mask, all of which were generated during the $\chi$-separation processing.

**[Step 1: Seed generation]**

In this step, a seed map for vessels was generated. Seeds for large and small vessels were obtained separately and then combined to form the final seed map. This approach improved identification of small vessels. The $R_2^*$, $\chi_{para}$, and $|\chi_{dia}|$ maps, which exhibited high signal intensity in vessels (see Supplementary Fig. 1), were utilized as the inputs for the seed generation. The $R_2^*$ map was exploited for large vessels whereas the product of the two susceptibility maps ($\chi_{para} \cdot |\chi_{dia}|$) was applied to identify small vessels.

For the large vessel seed map, a high pass filter (inverse Hamming filter; see Appendix A) was first applied to the $R_2^*$ map, suppressing large non-vessel structures of high $R_2^*$ (e.g., basal ganglia).[44,45] To remove high frequency residuals at the boundary of the brain, the $R_2^*$ map was inpainted outside the brain mask using coherent voxel values (*inpaintCoherent,* MATLAB) before the high pass filtering.[46] From this non-vessel suppressed $R_2^*$ map, a "vesselness" ($v_{MFAT}$) map was calculated by applying a multi-scale fractional anisotropy tensor (MFAT) filter (see Appendix B).[41] Finally, a threshold with a high cut-off value ($= mean(v_{MFAT}) + 2 \cdot std(v_{MFAT})$) was applied to the vesselness map, creating the seed map for large vessels.

For small vessels, maximum intensity projection (MIP) was applied to $\chi_{para} \cdot |\chi_{dia}|$ to enhance the visibility of small vessels, creating $\text{MIP}_{\chi_{para} \cdot |\chi_{dia}|}$.[37,47] The process was conducted every 16 mm, with half of the slices overlapping. During the MIP process, voxel positions corresponding to the maximum intensity position of $\chi_{para} \cdot |\chi_{dia}|$ were stored. To retain only small vessels, the large vessel seeds were removed from $\text{MIP}_{\chi_{para} \cdot |\chi_{dia}|}$ by negating the projection of large vessel seeds at the stored positions (1 - $\text{MIP}_{seed}$) and then multiplying it to $\text{MIP}_{\chi_{para} \cdot |\chi_{dia}|}$. From this result, a vesselness map was calculated using the MFAT filter and then the map was binarized with a low threshold ($= mean(v_{MFAT}) + 1 \cdot std(v_{MFAT})$). These seeds were back-projected to their original locations in 3D using the stored voxel positions, creating the small vessel seed map.

Finally, the large and small vessel seeds were combined to generate a final seed map.

**[Step 2: Vessel geometry guided-region growing]**

In this step, two vessel masks, one for $\chi_{para}$ and the other for $|\chi_{dia}|$, were generated through region growing guided by the geometry of vessels. Both masks were initialized with the final seed map from Step 1.

The region growing process began with defining a queue which stores all the seed voxels of the final seed map. The queue was re-ordered by the size of the seed cluster, prioritizing the largest seed cluster (*bwconncomp,* MATLAB) within which the smallest linear index of the voxel was queued first (*sub2ind,* MATLAB). Using this queue, the region growing algorithm ran as follows: The first element in the queue was selected as the starting seed voxel and removed from the queue. If an adjacent voxel of this seed voxel, which was not included in the vessel mask, met a region growing condition (see below), it was added to the queue and to the mask. The region growing process ended when the queue was empty.

The region growing condition is a criterion to include a voxel into the vessel mask. It has two susceptibility intensity limits:

$$upper\ limit = mean(\chi(seed = 1)) + \gamma_1 \cdot std(\chi(seed = 1)), \qquad \text{(Eq. 1)}$$

$$lower\ limit = mean(\chi(seed = 1)) + \gamma_2 \cdot std(\chi(seed = 1)), \qquad \text{(Eq. 2)}$$

where $\chi$ is either $\chi_{para}$ or $|\chi_{dia}|$ map, $seed$ is the final seed map, and $\gamma_1$ and $\gamma_2$ are hyperparameters to determine the intensity limits. If an adjacent voxel had an intensity higher than the upper limit, the voxel was added to the vessel mask. If the intensity of an adjacent voxel was lower than the lower limit, the voxel was ignored. If the intensity was between the upper and lower limits, the voxel was incorporated into the mask when it satisfied Eq. 3, which is composed of directionality similarity, intensity similarity, and anisotropy:

$$v_{MFAT}(q) \geq 0.5 \cdot \frac{1-\Omega(p,q)}{R(p,q)\cdot(1-e^{-10\cdot Ani(q)})}, \qquad \text{(Eq. 3)}$$

where

$$\Omega(p,q) = \frac{v_1(p)\cdot v_1(q)}{\|v_1(p)\|\|v_1(q)\|}, \qquad \text{(Eq. 4 Directionality similarity)}$$

$$R(p,q) = \begin{cases} \frac{I(p)}{I(q)}, & if\ I(p) \leq I(q), \\ \frac{I(q)}{I(p)}, & if\ I(p) > I(q), \end{cases} \qquad \text{(Eq. 5 Intensity similarity)}$$

$$Ani(q) = |\lambda_2(q) \cdot \lambda_3(q)|, \qquad \text{(Eq. 6 Anisotropy)}$$

with $p$ is the location of the seed voxel, $q$ is the location of an adjacent voxel, $v_{MFAT}$ is vesselness determined by the MFAT filter from the susceptibility map (either $\chi_{para}$ or $|\chi_{dia}|$), and $I$ denotes the intensity of the image. $v_1$ is an eigenvector corresponding to the smallest eigenvalue $\lambda_1$ of the Hessian matrix of the susceptibility map, indicating the direction of the vessel. $\lambda_2$ and $\lambda_3$ are the other two eigenvalues, which characterize variances in the directions perpendicular to the vessel direction. Eq. 3 was modified from Kerkeni et al., which combined vesselness and directional information of vessels (i.e., directionality similarity).[48] The two new criteria, intensity similarity and anisotropy, were designed to remove non-vessel structures effectively in Step 3 (see Supplementary Fig. 2). The anisotropy criterion was formulated in exponential form to express it as a probability value in Eq. 3 as suggested by Frangi et al.[38]

This vessel geometry guided-region growing generated initial vessel masks, one for $\chi_{para}$ and the other for $|\chi_{dia}|$.

**[Step 3: Non-vessel structure removal]**

In this step, the initial vessel masks were refined by excluding non-vessel structures, generating a final vessel mask.

In the initial vessel mask, non-vessel structures such as globus pallidus could also be included because of their high susceptibility intensity (see $CC_2$ and $CC_3$ in Fig. 1). To remove these structures, the structural characteristic of vessels, which reported a high anisotropy value, was exploited. The initial vessel mask was clustered into connected components (CCs) and then CCs with low anisotropy were removed as follows:

$$\frac{1}{N}\sum_{p \in CC_n} |\lambda_2(p) \cdot \lambda_3(p)| < Aniso\_Thresh, \qquad \text{(Eq. 7)}$$

where $N$ indicates the number of voxels in a CC, $CC_n$ denotes the $n^{th}$ CC, $Aniso\_Thresh$ is an anisotropy threshold, and $|\lambda_2(p) \cdot \lambda_3(p)|$ is anisotropy criterion defined in Step 2.

The final vessel mask was generated by accumulating the remaining CCs and binarizing it. Two masks, one for $\chi_{para}$ and the other for $|\chi_{dia}|$, were produced.

## 2.2 MRI data acquisition and data processing

In this study, three datasets from previous studies were used: the $\chi$-sepnet dataset,[10,49] the $\chi$-separation template dataset,[27] and the high-resolution $\chi$-separation dataset.[50] The study was approved by the institutional review board.

The $\chi$-sepnet dataset is from 12 subjects and is composed of 3D multi-echo GRE data in six head orientations and one 2D multi-echo spin echo (MESE) data (3T, Siemens Tim Trio, Erlangen, Germany). The $\chi$-separation template dataset has 106 subjects, with 3D multi-echo GRE images and T1-weighted images obtained using MPRAGE (3T, Philips Ingenia CX and Ingenia Elition X, Amsterdam, Netherlands). In the high-resolution $\chi$-separation dataset, 3D multi-echo GRE data from 8 subjects were utilized (7T, Siemens Magnetom Terra, Erlangen, Germany). The acquisition parameters of the three datasets are summarized in Supplementary Table 1.

For all three datasets, a common data processing pipeline was applied: A brain mask was generated from the first echo magnitude image of the multi-echo GRE data, using BET (FSL, FMRIB, Oxford, UK).[51] Phase processing for a local field map followed the QSM consensus guideline.[52] Briefly, phase images from the multi-echo GRE data were unwrapped using the rapid opensource minimum spanning tree algorithm (ROMEO).[53] The unwrapped phase images were averaged using a weighted echo sum to produce a combined phase image.[54] Then, background field removal using V-SHARP was applied, creating a local field map.[55,56] From the multi-echo GRE magnitude images, an $R_2^*$ map was generated by voxel-wise fitting of a mono-exponential decay function using a nonlinear least square solver (*lsqnonlin*, MATLAB). For $R_2$ mapping, a simulated dictionary of spin-echo decay, constructed with the StimFit toolbox,[57,58] was utilized to match for the MESE magnitude images.

For the $\chi$-sepnet dataset, the local field and $R_2^*$ maps from each orientation, as well as the $R_2$ map, were registered to the first head orientation using FSL FLIRT.[59] A registration matrix was computed from the first echo magnitude images of GRE and MESE, and subsequently applied to align the local field, $R_2^*$, and $R_2$ maps. The $R_2'$ map was then generated by subtracting the registered $R_2$ map from each orientation $R_2^*$ map, with negative values set to zero. Finally, $\chi_{para}$ and $|\chi_{dia}|$ maps were generated via four $\chi$-separation algorithms: $\chi$-sep-COSMOS,[60] $\chi$-sep-MEDI,[9] $\chi$-sep-iLSQR,[9] and $\chi$-sepnet-$R_2^*$.[49] For $\chi$-sep-COSMOS, which is a multi-orientation $\chi$-separation algorithm, the local field and $R_2'$ maps from all orientations were utilized. For $\chi$-sep-MEDI and $\chi$-sep-iLSQR, which are single-orientation conventional $\chi$-separation algorithms, the local field and $R_2'$ maps from the first orientation were used as the input. Lastly, for $\chi$-sepnet-$R_2^*$, a deep learning-based $\chi$-separation algorithm designed for single-orientation GRE data only, the local field and $R_2^*$ maps from the first orientation were used. For the $\chi$-separation template dataset, $\chi_{para}$ and $|\chi_{dia}|$ maps were generated using $\chi$-sepnet-$R_2^*$. For the high-resolution $\chi$-separation dataset, the pipeline proposed by J. Kim et al. was applied to produce high-resolution $\chi$-separation maps.[50]

Finally, the proposed vessel segmentation method was applied to all the $\chi$-separation maps. Out of the hyperparameters, $\gamma_1$ and $\gamma_2$ were fixed as 0.5 and -0.5, respectively. On the other hand, the anisotropy threshold ($Aniso\_Thresh$) was adjusted for each subject with $1.2 \times 10^{-3}$ as the starting point for both $\chi_{para}$ and $|\chi_{dia}|$ maps. In most cases, this starting point value created a high-quality outcome, however, in some cases, the value was increased when deep gray matter regions were not properly excluded.

**2.3** Comparison with conventional vessel segmentation methods

The proposed vessel segmentation method was compared with two previously proposed methods: the Frangi filter[38] and a GRE-based vessel segmentation method.[44] For the Frangi filter, a $\chi_{para}$ (or $|\chi_{dia}|$) map was used as input. The Frangi filter had six parameters: scale range ($\sigma$), scale ratio ($\Delta\sigma$), Frangi vesselness constants ($\alpha$, $\beta$, and $c$), and a threshold.[38] Based on previous studies,[36,44] the parameters were set as follows: $\sigma = [0.25, 2.5]$, $\Delta\sigma = 0.25$, $\alpha = 0.5$, $\beta = 0.5$, $c =$ half of the maximal Hessian norm, and threshold $= 0.02$. The Frangi filter was publicly available (https://kr.mathworks.com/matlabcentral/fileexchange/24409-hessian-based-frangi-vesselness-filter). The GRE-based vessel segmentation method was originally developed to extract veins by leveraging QSM, SWI, and $R_2^*$. To adapt this method for $\chi$-separation, QSM was replaced by the $\chi_{para}$ (or $|\chi_{dia}|$) map. This method was available online (https://github.com/SinaStraub/GRE_vessel_seg).

For the evaluation of the performances, three subjects from the $\chi$-sepnet dataset (3T) and three subjects from the high-resolution $\chi$-separation dataset (7T) were manually segmented using ITK-snap[61] to produce ground truth segmentation results (see Supplementary Fig. 3). This manual segmentation was conducted on central 12 consecutive slices on the axial, sagittal, and coronal planes for each subject (total 36 slices per subject), including deep gray matter regions and large and small cerebral vessels. For quantitative assessment, the Dice similarity coefficient (DSC) was calculated.[62] Additionally, processing time and memory usage of each method were evaluated.

All methods were executed on a workstation with an Intel Xeon CPU E5-2699 v4 @ 2.20 GHz and 396 GB RAM.

**2.4** Robustness of the proposed vessel segmentation method

To assess the robustness of the proposed vessel segmentation method, the method was applied to $\chi$-separation maps from the four $\chi$-separation algorithms ($\chi$-sep-COSMOS, $\chi$-sep-MEDI, $\chi$-sep-iLSQR, and $\chi$-sepnet-$R_2^*$). The three 3T subject data with the manual segmentation masks were utilized to calculate DSC for the vessel mask generated from the $\chi$-separation map of each algorithm.

### 2.5 Applications of the proposed vessel segmentation method

To demonstrate the utility of vessel masks when analyzing $\chi$-separation maps, vessel masks were applied to two applications: quantitative evaluation of $\chi$-sepnet-$R_2^*$ and population-averaged region of interest (ROI) analysis.

To evaluate the effects of vessels in assessing the reconstruction performance of $\chi$-sepnet-$R_2^*$, the root mean squared error (RMSE), peak signal-to-noise ratio (PSNR), and structure similarity index (SSIM) were calculated under three conditions: including vessels (without applying the vessel mask), excluding vessels (with the vessel mask), and within the vessel mask. Each metric was computed with respect to $\chi$-sep-COSMOS as the reference. The $\chi$-sepnet-$R_2^*$ trained in M. Kim et al. was utilized,[49] and the evaluation was conducted using the test data from the $\chi$-sepnet dataset (six subjects with the six head orientations).

To assess the impact of vessels on the population-averaged ROI analysis, the proportion of vessels and the population average of the mean susceptibility values with and without vessels were quantified across twenty-seven ROIs defined in the $\chi$-separation atlas.[27] Data from 106 subjects in the $\chi$-separation template dataset were utilized for this analysis, and the ROIs were transformed into each subject's space using deformation matrices in Min et al.[27] The proportion of vessels in each ROI was calculated as the ratio of the number of voxels included in the vessel mask to the total number of voxels in the ROI. The population average of mean susceptibility values for each ROI was computed with and without vessels. A paired *t*-test was conducted to discern statistically significant differences between two measurements, with significance determined at a Bonferroni-corrected threshold of $p < 0.05$.

# 3. Results

Vessel segmentation results of the two conventional methods and the proposed method are illustrated in Fig. 2 for $\chi_{para}$ and Fig. 3 for $|\chi_{dia}|$. The proposed method demonstrates superior performance by effectively excluding non-vessel structures (yellow arrows in Figs. 2 and 3: globus pallidus in $\chi_{para}$ and optic radiation in $|\chi_{dia}|$) while maintaining sensitivity to vessels. Furthermore, the conventional methods fail to capture parts of vessels (green arrows in Figs. 2 and 3), while the proposed method achieves clear masking of vessels. In a few small vessels, however, the GRE-based vessel segmentation method results in better outcomes than the proposed method (blue arrows in Figs. 2 and 3). The superior performance of the proposed method is consistent across the resolutions of 1 × 1 × 1 mm³ (3T) and 0.65 × 0.65 × 0.65 mm³ (7T), which can be confirmed in the MIP images of Fig. 4. The quantitative metrics in Table 1 consolidate that the proposed method achieves the best performance, reporting the highest DSCs.

The proposed method significantly reduces both processing time and memory usage compared to the GRE-based vessel segmentation method. For the 3T data (matrix size: 256 × 224 × 176), the proposed method requires only 2 GB of RAM and takes 4 minutes for processing, while the GRE-based vessel segmentation method demands approximately 60 GB of RAM and 28 minutes for processing. For the 7T high-resolution data (matrix size: 350 × 284 × 224), the proposed method needs 4 GB of RAM and 16 minutes, whereas the GRE-based vessel segmentation method requires 140 GB of RAM and 80 minutes. The Frangi filter takes 76 seconds with 1.2 GB of RAM for processing the 3T data and 3 minutes with 2 GB of RAM for processing the 7T data.

The proposed vessel segmentation method demonstrates robust performance across the four $\chi$-separation algorithms (Fig. 5), showing consistent segmentation results for both $\chi_{para}$ and $|\chi_{dia}|$. Minor differences are observed due to the characteristics of each algorithm, particularly in cortical areas (yellow arrows in Fig. 5; see Discussion). The DSC scores (Table 2) also confirm comparable segmentation performance across the four $\chi$-separation algorithms.

When the vessel mask was applied to the performance evaluation of $\chi$-sepnet-$R_2^*$ against $\chi$-sep-COSMOS, the quantitative metrics (RMSE, PSNR, and SSIM) report notable improvements (Table 3), demonstrating a potential value of masking out vessels if they are not of interest. Within the vessel mask, we observed higher RMSE and lower PSNR and SSIM than the non-vessel regions (i.e., with vessel mask results).

In the evaluation of the population-averaged ROI values of $\chi_{para}$ and $|\chi_{dia}|$, the vessel mask does make statistically significant differences in the susceptibility values in 16 out of 27 ROIs (Table 4). For example, $\chi_{para}$ of caudate, which reports the highest vessel proportion (3.76 ± 1.67 %) due to the inclusion of the anterior terminal vein (Fig. 6), shows a statistically significant difference in the susceptibility value (47.5 ± 7.2 ppb without the mask vs. 44.4 ± 6.8 ppb with the mask). In $|\chi_{dia}|$, statistically significant differences are observed in most ROIs, with the genu of the corpus callosum showing the greatest reduction in the population-averaged mean susceptibility values (32.2 ± 3.0 ppb without the mask vs. 30.7 ± 2.9 ppb with the mask). The genu ROI is primarily influenced by the septal vein, requiring exclusion of the vessel (Fig. 6). These examples suggest the importance of the vessel mask for accurate estimation of the ROI susceptibility values.

## 4. Discussion

In our study, a new vessel segmentation method is developed to enhance the quantification of iron and myelin content in $\chi$-separation. Our results demonstrated superior performance of the proposed method compared to the conventional approaches by effectively excluding non-vessel structures, achieving the highest DSC.

Our method has three hyperparameters ($\gamma_1$, $\gamma_2$, and anisotropy threshold) that can be adjusted for segmentation quality. Among them, $\gamma_1$ and $\gamma_2$ were fixed in this study whereas the anisotropy threshold was optimized for each subject. The choice of the anisotropy threshold is a balance between excluding non-vessel structures and including small vessels. When tested on the $\chi$-separation maps from $\chi$-sep-COSMOS and $\chi$-sepnet-$R_2^*$, a typical anisotropy threshold value ranged from 0.0012 to 0.0048 for $\chi_{para}$ and from 0.0012 to 0.0024 for $|\chi_{dia}|$. It was higher for $\chi$-sep-MEDI and $\chi$-sep-iLSQR maps (from 0.0072 to 0.0108 for $\chi_{para}$ and from 0.0018 to 0.0048 for $|\chi_{dia}|$), which might have led to the exclusion of small vessels (yellow arrows in Fig. 5) and resulted in slightly reduced DSC values (Table 2).

The proposed method also allows to adjust the $\gamma$ parameters as optional hyperparameters to further improve performance. As suggested in Eq. 1, $\gamma_1$ controls the inclusion of voxels with high intensity, affecting sensitivity to vessels. A low $\gamma_1$ value enhances the delineation of vessels but may lead to the inclusion of non-vessel structures with high intensity. In our study, the default settings ($\gamma_1$ = 0.5) worked well in most cases. In some cases, however, falsely included deep gray matter regions, due to a low $\gamma_1$, was connected to nearby large vessels (see Supplementary Fig. 4). Such outcomes tended to occur in datasets from old populations, where increased susceptibility values in deep gray matter regions are observed due to iron accumulation.[63,64] By employing a higher $\gamma_1$ value, these masks can be improved. Additionally, for $|\chi_{dia}|$ maps, meninges, which reveal high absolute susceptibility values, were prone to be included in the vessel masks when $\gamma_1$ was low.

We observed that the conventional methods often included non-vessel structures such as deep gray matter and highly myelinated fibers (see Figs. 2, 3, and 4). The Frangi filter has adjustable parameters, which was set to the values of previous QSM studies.[36,44] When it was adjusted for individual subjects, performance improved but the results still faced a trade-off between sensitivity to vessels and the inclusion of non-vessel structures (see Supplementary Figs. 5 and 6). The GRE-based vessel segmentation method improved the exclusion of non-vessels by leveraging additional contrasts (SWI and $R_2^*$) but had no adjustable parameter. This approach, however, still failed to completely exclude non-vessels.

Our method has three challenges. First, subject-wise parameter tuning can be challenging for a large number of subjects. Second, the method struggles to differentiate calcification from vessels due to their hyperintensities as well as their tube-like structures in some cases (see Supplementary Fig. 3). Lastly, as previously mentioned, some small vessels were removed for the fixed parameters.

Deep learning has been widely applied to segmentation,[65–67] including brain vessel segmentation.[68–70] These approaches adopt data-driven manners to generate a vessel mask. By learning structural information, deep learning models have potential to improve segmentation performance. However, generation of a training dataset requires high-quality vessel masks. Vessel masks obtained using the proposed method, combined with individually tuned hyperparameters and manual refinements, could serve as a high-quality label for training. This holds the potential to develop a method that is less reliant on hyperparameters. Nonetheless, generalization, such as resolution or contrasts, remains as critical consideration.[68,69,71]

Another progress in deep learning is foundation models for universal segmentation, such as MedSAM.[72] These models aim to generalize across a wide range of segmentation tasks. However, foundation models experience performance degradation on less-represented contrasts, necessitating fine-tuning.[72,73] Additionally, they have reported difficulties in segmenting vessel-like branching structures.[72] This limitation suggests that further research is required to adapt these models effectively for vessel segmentation tasks.

As for future work, our proposed method can be generalized for other susceptibility imaging contrasts. SWI and QSM, for instance, are potential.[28,30,31] By modifying the inputs of the method (i.e., $R_2^*$ and the product of $\chi_{para}$ and $|\chi_{dia}|$) to the corresponding contrasts (i.e., SWI and QSM) and adjusting the hyperparameters, the proposed method may achieve vessel segmentation for the contrasts.

# 5. Conclusion

This study proposes a vessel segmentation method for $\chi$-separation, leveraging region growing guided by vessel geometry. The vessel mask from this method outperforms the conventional methods, effectively excluding the non-vessel structures such as deep gray matters in $\chi_{para}$ and myelinated fibers in $|\chi_{dia}|$. We demonstrate that the method generates robust results across different resolutions and $\chi$-separation algorithms. Finally, the utility of the vessel masks in the analysis of $\chi$-separation maps suggest improvements in the reliability and accuracy of the analysis.

# Appendix

*A. Inverse Hamming filter*

An inverse Hamming filter, which is a high pass filter for suppressing non-vessel structures and enhancing vascular structures,[44,45] is formulated as follows:

$$iH(k_x, k_y, k_z) = \begin{cases} 0.6 \cdot \left[1 - \cos\left(\pi \sqrt{\frac{k_x^2}{H_x^2} + \frac{k_y^2}{H_y^2} + \frac{k_z^2}{H_z^2}}\right)\right], & \text{if } \frac{k_x^2}{H_x^2} + \frac{k_y^2}{H_y^2} + \frac{k_z^2}{H_z^2} \leq 1, \\ 1, & \text{otherwise,} \end{cases} \quad \text{(Eq. A1)}$$

where $k_x$, $k_y$, and $k_z$ are k-space index, and $H_x$, $H_y$, and $H_z$ are filter sizes in each axis. In this study, the filter sizes suggested by Straub et al.[44] ($H_x = H_y = H_z = 80$) were utilized.

*B. Multi-scale fractional anisotropy tensor vesselness ($v_{MFAT}$)*

Multi-scale fractional anisotropy tensor vesselness ($v_{MFAT}$) reports the likelihood of a voxel being vascular. This $v_{MFAT}$ is an extension of fractional anisotropy tensor vesselness ($v_{FAT}$), effectively identifying vessels of varying sizes.[41]

$v_{FAT}$ is derived from the eigenvalues ($\lambda_1, \lambda_2, \text{ and } \lambda_3$; $|\lambda_1| < |\lambda_2| < |\lambda_3|$) of a Hessian matrix of an input image[41]:

$$v_{FAT} = \sqrt{\frac{3}{2} \cdot \frac{(\lambda_2 - \bar{D}_\lambda)^2 + (\lambda_\rho - \bar{D}_\lambda)^2 + (\lambda_v - \bar{D}_\lambda)^2}{(\lambda_2)^2 + (\lambda_\rho)^2 + (\lambda_v)^2}}, \quad \text{(Eq. A2)}$$

where $\bar{D}_\lambda$ is the mean of the eigenvalues ($= \frac{\lambda_1 + \lambda_2 + \lambda_3}{3}$), and $\lambda_\rho$ and $\lambda_v$ are

$$\lambda_{\rho \text{ or } v} = \begin{cases} \lambda_3, & \text{if } \lambda_3 < \tau_{\rho \text{ or } v} \cdot \min_{\mathbf{r}} \lambda_3, \\ \tau_{\rho \text{ or } v} \cdot \min_{\mathbf{r}} \lambda_3, & \text{if } \tau_{\rho \text{ or } v} \cdot \min_{\mathbf{r}} \lambda_3 \leq \lambda_3 < 0, \\ 0, & \text{otherwise,} \end{cases} \quad \text{(Eq. A3)}$$

where $\mathbf{r}$ is the voxel position, and $\tau_\rho$ and $\tau_v$ are scaling constants between 0 and 1, modifying the eigenvalues to ensure the robustness of vesselness to low magnitudes of $\lambda_2$ and $\lambda_3$.

To consider various sizes of vessels, $v_{FAT}$ is calculated for input images filtered by Gaussian kernels with different standard deviations $\sigma = \{\sigma_1, \sigma_2, \cdots, \sigma_m\}$. Then, $v_{MFAT}$ is obtained as follows[41]:

$$v_{MFAT}^{\sigma_1} = R_\lambda^{\sigma_1}, \quad \text{(Eq. A4)}$$

$$v_{MFAT}^{\sigma_j} = v_{MFAT}^{\sigma_{j-1}} + \delta \cdot \tanh\left(R_\lambda^{\sigma_j} - \delta\right), \quad \text{(Eq. A5)}$$

$$v_{MFAT}^{\sigma_j} = max(v_{MFAT}^{\sigma_j}, R_\lambda^{\sigma_j}), \tag{Eq. A6}$$

where $\sigma_j$ is the current standard deviation, $\sigma_{j-1}$ is a previous standard deviation, $\delta$ is the step size, and $R_\lambda^{\sigma_j}$ is defined as:

$$R_\lambda^{\sigma_j} = \begin{cases} 0, & if\ \lambda_\rho > \lambda_\rho - \lambda_2\ or\ \lambda_\rho \geq 0\ or\ \lambda_2 \geq 0, \\ 1, & if\ \lambda_\rho - \lambda_2 = \max_r(\lambda_\rho - \lambda_2), \\ 1 - v_{FAT}^{\sigma_j}, & otherwise. \end{cases} \tag{Eq. A7}$$

In this study, the default parameters in Alhasson et al.[41] were used ($\sigma = [0.25, 1]$ with an increment ($\Delta\sigma$) of $0.25$, $\tau_\rho = 0.02$, $\tau_\nu = 0.35$, and $\delta = 0.3$). The implementation of MFAT is available online (https://github.com /Haifafh/MFAT).

## Data availability statement

The proposed method is implemented in MATLAB and is available as a part of the $\chi$-separation toolbox (https://github.com/SNU-LIST/chi-separation). The data used in this study are available on request from the corresponding author and can be shared following Institutional Review Board approval due to privacy or ethical restrictions.

## Funding Information

This work was supported by National Research Foundation of Korea (NRF) (RS-2024-00349509), Korea Health Industry Development Institute (KHIDI) (RS-2024-00439677), Korea Basic Science Institute (RS-2024-00435727), Institute of Information & communications Technology Planning & Evaluation (IITP) (IITP-2023-RS-2023-00256081), and Institute of New Media and Communications, Institute of Engineering Research, and AI-BIO Research Grant of Seoul National University.

# Figures

**Figure 1.** Overview of the proposed pipeline for vessel segmentation. The pipeline has three steps: Step 1 for seed generation, Step 2 for an initial vessel mask created by region growing guided by the characteristics of vessel geometry, and Step 3 for non-vessel structures removal.

**Figure 2.** Results of the vessel segmentation methods applied to $\chi_{para}$. The $\chi_{para}$ maps (first column) and the three vessel segmentation masks overlaid on $\chi_{para}$ (second column: Frangi filter, third column: GRE-based method, and fourth column: proposed method) are displayed. Three representative slices that include the globus pallidus (first row), a large vein (second row), and small vessels (third row) reveal that the proposed method effectively excludes non-vessel structures (yellow arrows), providing a high-quality vessel mask (green arrows). For small vessels, however, the GRE-based method shows more sensitivity (blue arrows).

**Figure 3.** Results of the vessel segmentation methods applied to $|\chi_{dia}|$. The $|\chi_{dia}|$ maps (first column) and the three vessel segmentation masks overlaid on $|\chi_{dia}|$ (second column: Frangi filter, third column: GRE-based method, and fourth column: proposed method) are displayed. Three representative slices that include the optic radiation (first row), cortical vessels (second row), and small vessels (third row) reveal that the proposed method effectively excludes non-vessel structures (yellow arrows), providing a high-quality vessel mask (green arrows). For small vessels, however, the GRE-based method shows more sensitivity (blue arrows).

**Figure 4.** MIP of $\chi_{para}$ and $|\chi_{dia}|$ maps, and vessel segmentation outcomes from 3T data (upper rows, $1 \times 1 \times 1$ mm³ resolution) and 7T data (lower rows, $0.65 \times 0.65 \times 0.65$ mm³ resolution). The conventional methods erroneously segment deep gray matter structures (yellow arrows) or miss large vessels (green arrows) whereas the proposed method delivers more accurate results.

**Figure 5.** Vessel segmentation results from the four different $\chi$-separation algorithms: $\chi$-sep-COSMOS, $\chi$-sep-MEDI, $\chi$-sep-iLSQR, and $\chi$-sepnet-$R_2^*$. The results for (a) $\chi_{para}$ and (b) $|\chi_{dia}|$ demonstrate consistent segmentation of vessels across all algorithms, showing the robustness of the proposed method. Minor differences occur in small vessels within cortical regions (yellow arrows).

**Figure 6.** Representative ROIs including vessels. Caudate and corpus callosum show the highest vessel portion for $\chi_{para}$ and $|\chi_{dia}|$, respectively. Caudate primarily includes the anterior terminal veins whereas corpus callosum has septal veins (yellow arrows).

# Tables

**Table 1**. Quantitative comparison of the vessel segmentation methods.

**Table 2**. DSCs for the four different $\chi$-separation algorithms.

**Table 3.** Reconstruction quality of $\chi$-sepnet-$R_2^*$ with respect to $\chi$-sep-COSMOS when analyzed with, without, and within the vessel mask.

**Table 4.** Application for the population-averaged ROI analysis. This table reports the proportion of vessels, and the mean susceptibility in each ROI analyzed with and without vessel masks.

# Supporting Information

**Supplementary Figure 1.** Representative slice of $R_2^*$, $\chi_{para}$, $\chi_{dia}$ and $\chi_{para} \cdot |\chi_{dia}|$ showing high signal intensities within vessels across all maps (red arrows). Veins are typically paramagnetic whereas arteries possess susceptibility close to surrounding tissues. In $\chi$-separation, however, both veins and arteries exhibit inaccurate values within and near vessel regions because of flow artifacts and non-local R2* effects. Many of them appear on both susceptibility maps, creating vessel artifacts.

**Supplementary Figure 2.** Ablation study of the region growing conditions: (a) Kerkeni et al.'s condition without intensity limits. (b) Proposed condition without intensity limits. (c) Kerkeni et al.'s condition with intensity limits. (d) Proposed condition with intensity limits. When intensity limits were not used, the algorithm showed reduced sensitivity to large vessels (blue arrows). Additionally, the use of the proposed condition demonstrated its effectiveness in excluding non-vessel structures, such as basal ganglia (yellow arrows).

**Supplementary Figure 3.** Manual segmentation results for $\chi_{para}$ and $|\chi_{dia}|$ displayed in (a, d) axial, (b, e) sagittal, and (c, f) coronal views. Manual segmentation was conducted on 12 consecutive slices in the axial, sagittal, and coronal planes (a total of 36 slices per subject for each susceptibility map) using ITK-snap. Calcifications (yellow arrows), meninges (orange arrows), and artifacts caused by mis-registration between $R_2^*$ and $R_2$ (blue arrows), which exhibit high intensity in both $\chi_{para}$ and $|\chi_{dia}|$, were excluded from these masks.

**Supplementary Figure 4.** Effects of hyperparameters on vessel segmentation displayed on $\chi_{para}$ maps. (a) A lower $\gamma_1$ value results in clearer vessel delineation (yellow arrow). (b) A higher anisotropy threshold better excludes non-vessel structures, but may also exclude vessels (orange arrows). (c) The worst-case scenario, where either non-vessel structures are not excluded or both non-vessels and vessels are excluded together, can occur (blue box).

**Supplementary Figure 5.** ROC curves for optimizing the Frangi filter parameters for (a) 3T $\chi_{para}$, (b) 3T $|\chi_{dia}|$, (c) 7T $\chi_{para}$ and (d) 7T $|\chi_{dia}|$. The optimum parameters were selected that achieved the maximum specificity while exceeding the sensitivity of the proposed method.

**Supplementary Figure 6.** Comparison of vessel masks generated using the Frangi filter with the default and optimized parameters. Vessel masks generated with the optimized parameters show improved results, excluding more deep gray matter regions in $\chi_{para}$ (yellow arrows) and capturing more small vessels in $|\chi_{dia}|$ (orange arrows). The DSC values at the bottom of the zoomed-in images confirm the enhanced results for the optimized parameters. Despite the improvement, the vessel masks still include non-vessels structures, reporting lower DSC values than the proposed method.

**Supplementary Table 1**. Summary of the MRI acquisition parameters for the three datasets.

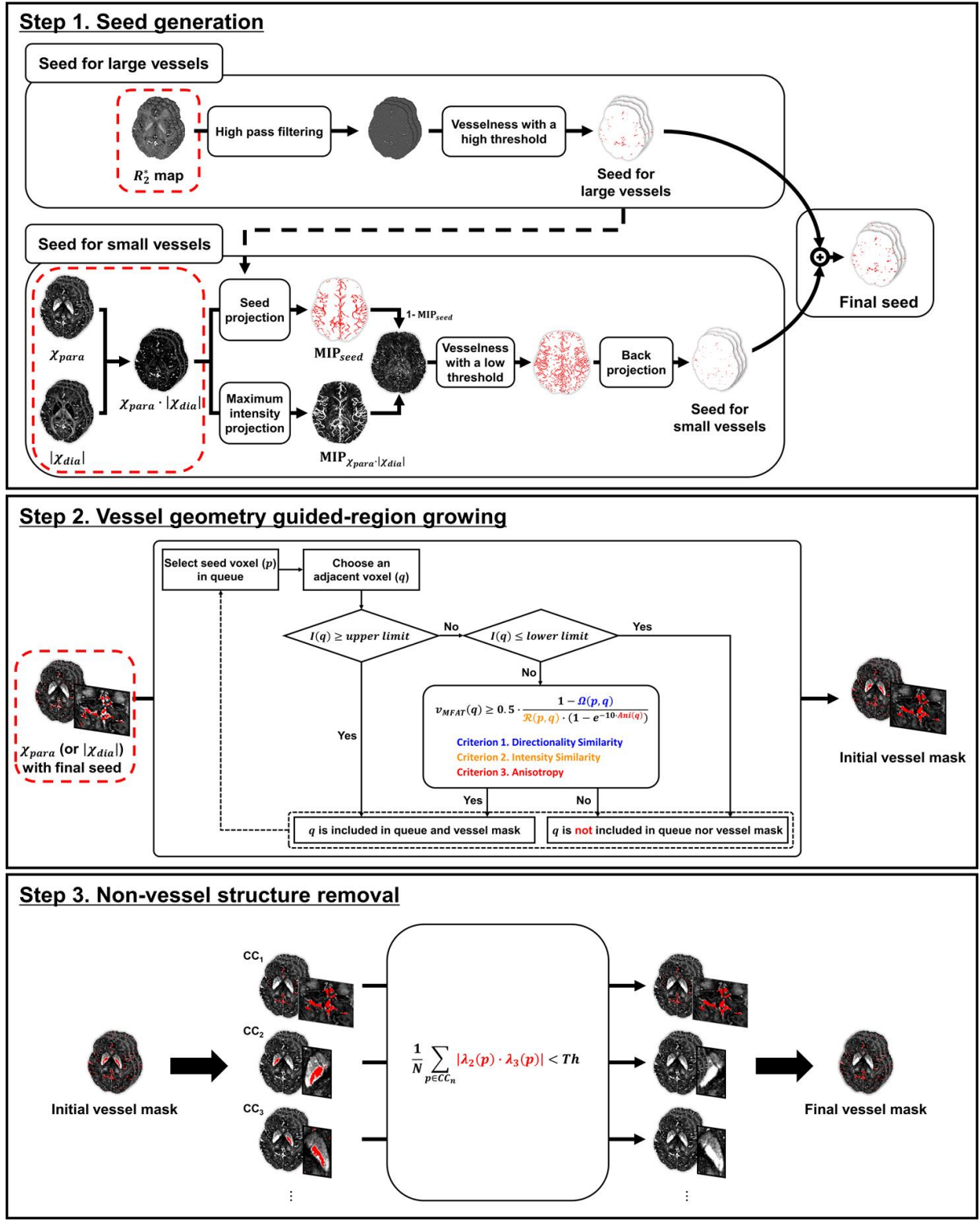

**Figure 1.** Overview of the proposed pipeline for vessel segmentation. The pipeline has three steps: Step 1 for seed generation, Step 2 for an initial vessel mask created by region growing guided by the characteristics of vessel geometry, and Step 3 for non-vessel structures removal.

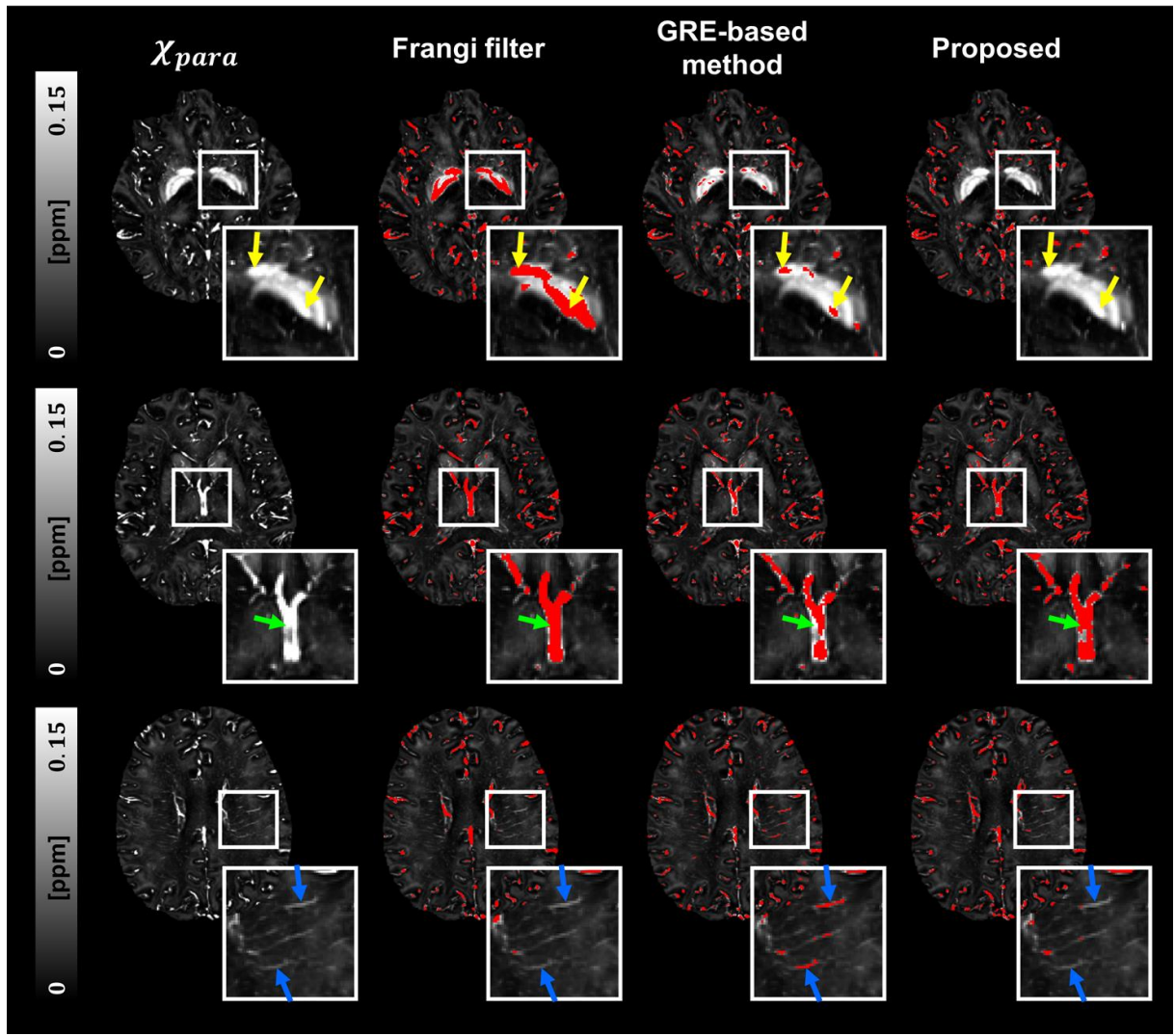

**Figure 2.** Results of the vessel segmentation methods applied to $\chi_{para}$. The $\chi_{para}$ maps (first column) and the three vessel segmentation masks overlaid on $\chi_{para}$ (second column: Frangi filter, third column: GRE-based method, and fourth column: proposed method) are displayed. Three representative slices that include the globus pallidus (first row), a large vein (second row), and small vessels (third row) reveal that the proposed method effectively excludes non-vessel structures (yellow arrows), providing a high-quality vessel mask (green arrows). For small vessels, however, the GRE-based method shows more sensitivity (blue arrows).

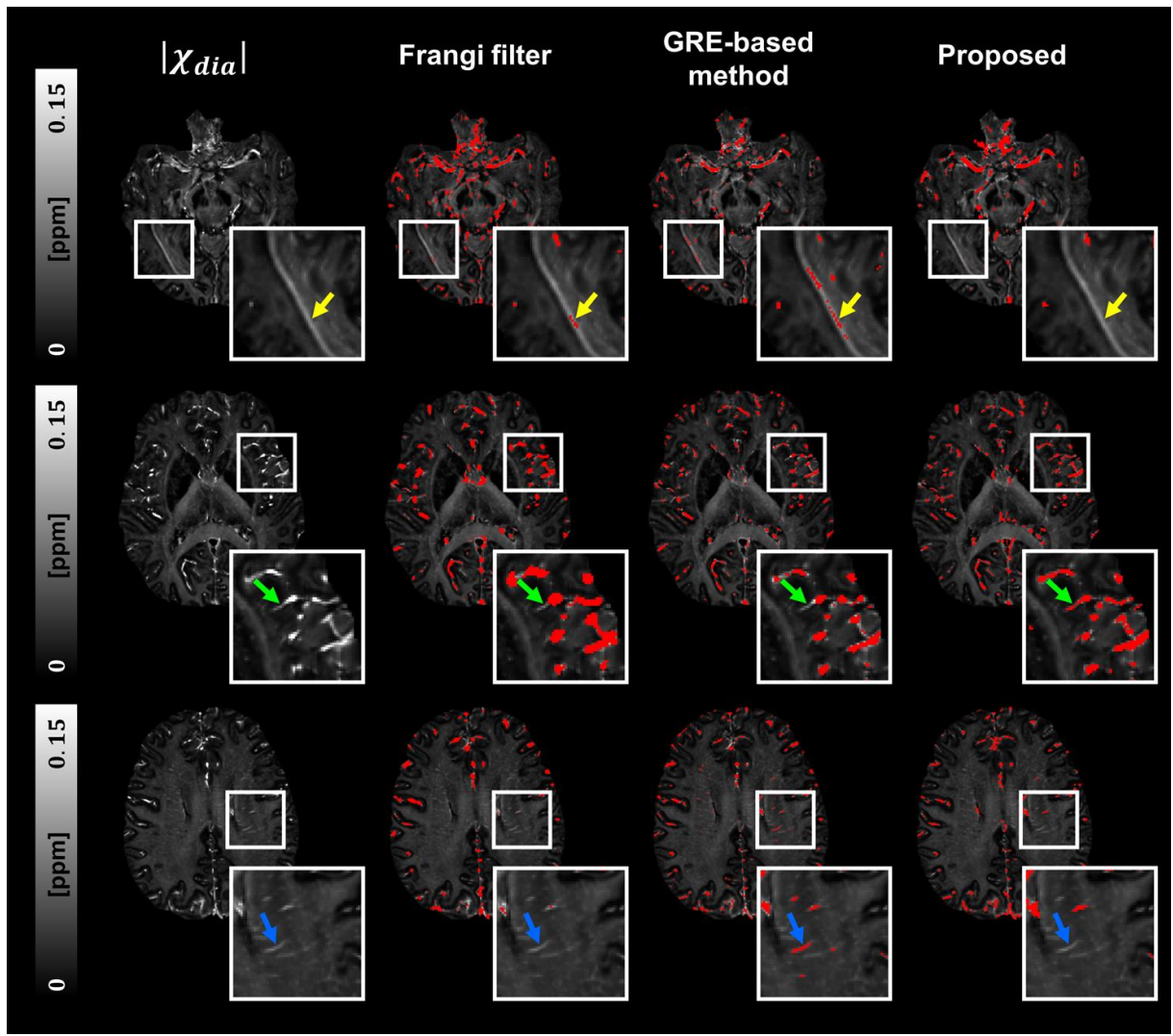

**Figure 3.** Results of the vessel segmentation methods applied to $|\chi_{dia}|$. The $|\chi_{dia}|$ maps (first column) and the three vessel segmentation masks overlaid on $|\chi_{dia}|$ (second column: Frangi filter, third column: GRE-based method, and fourth column: proposed method) are displayed. Three representative slices that include the optic radiation (first row), cortical vessels (second row), and small vessels (third row) reveal that the proposed method effectively excludes non-vessel structures (yellow arrows), providing a high-quality vessel mask (green arrows). For small vessels, however, the GRE-based method shows more sensitivity (blue arrows).

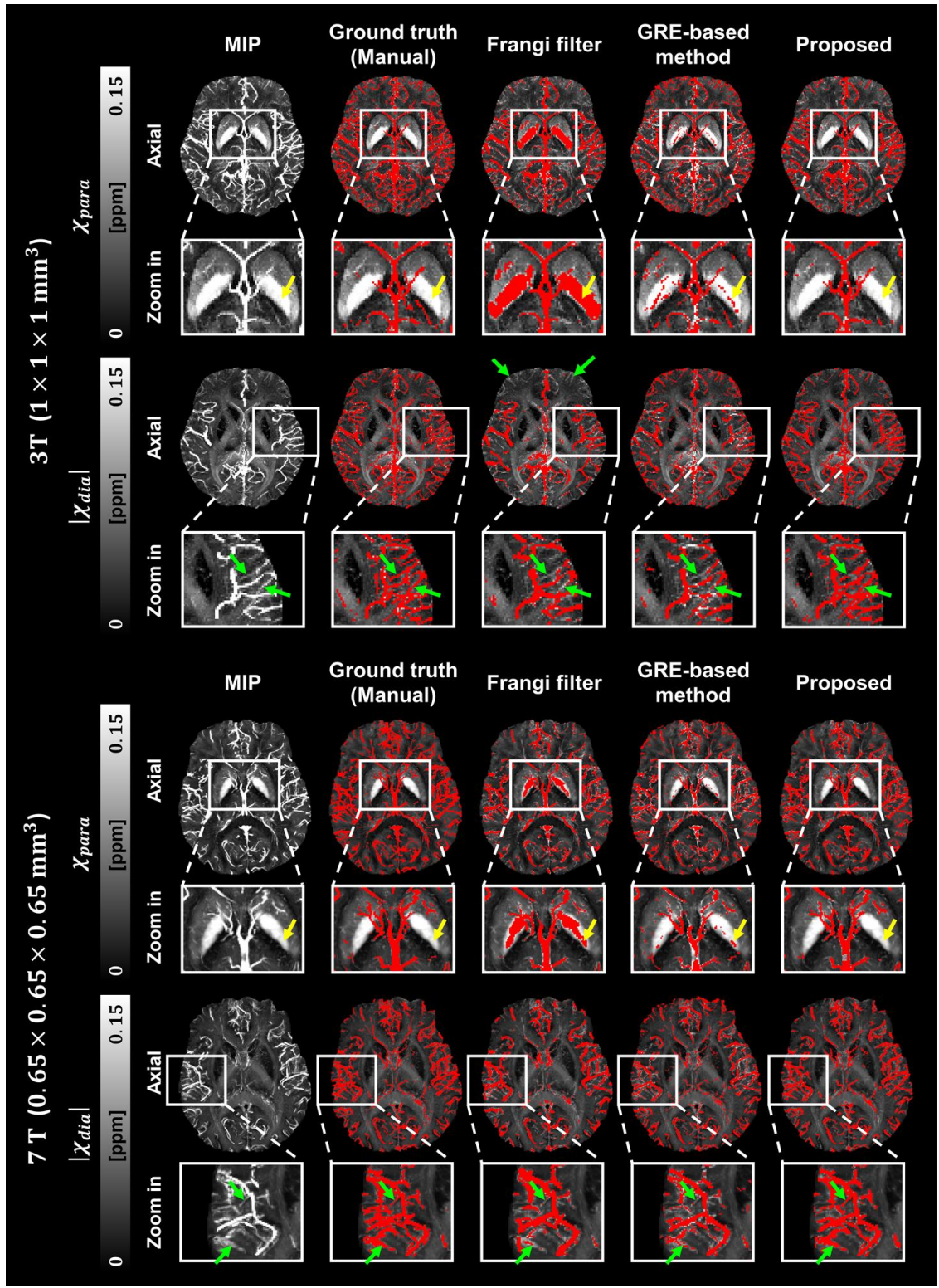

**Figure 4.** MIP of $\chi_{para}$ and $|\chi_{dia}|$ maps, and vessel segmentation outcomes from 3T data (upper rows, $1 \times 1 \times 1$ mm$^3$ resolution) and 7T data (lower rows, $0.65 \times 0.65 \times 0.65$ mm$^3$ resolution). The conventional methods erroneously segment deep gray matter structures (yellow arrows) or miss large vessels (green arrows) whereas the proposed method delivers more accurate results.

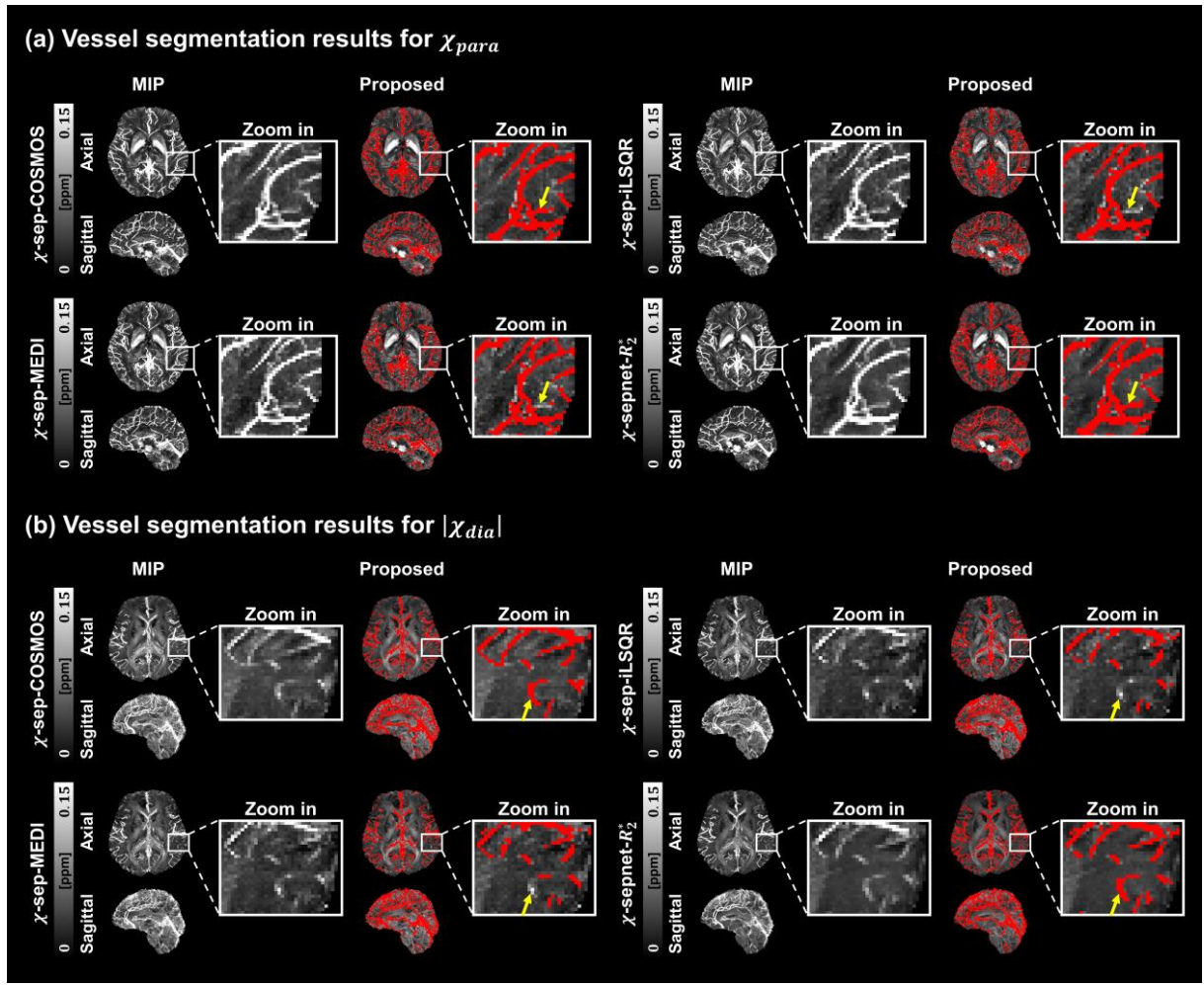

**Figure 5.** Vessel segmentation results from the four different $\chi$-separation algorithms: $\chi$-sep-COSMOS, $\chi$-sep-MEDI, $\chi$-sep-iLSQR, and $\chi$-sepnet-$R_2^*$. The results for (a) $\chi_{para}$ and (b) $|\chi_{dia}|$ demonstrate consistent segmentation of vessels across all algorithms, showing the robustness of the proposed method. Minor differences occur in small vessels within cortical regions (yellow arrows).

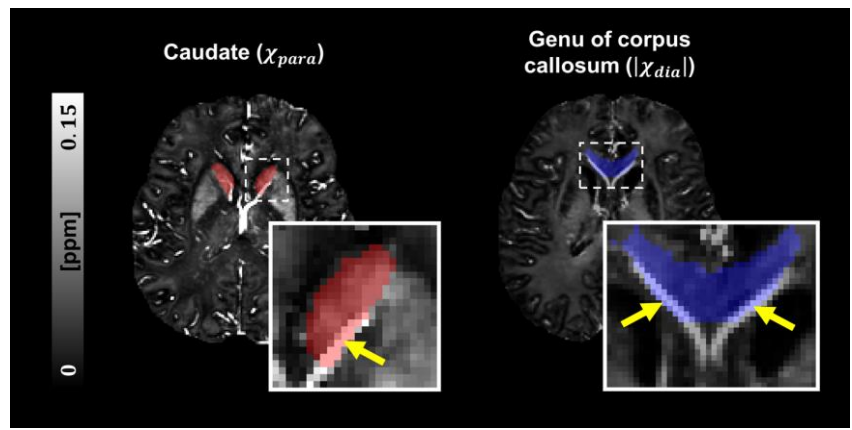

**Figure 6.** Representative ROIs including vessels. Caudate and corpus callosum show the highest vessel portion for $\chi_{para}$ and $|\chi_{dia}|$, respectively. Caudate primarily includes the anterior terminal veins whereas corpus callosum has septal veins (yellow arrows).

**Table 1**. Quantitative comparison of the vessel segmentation methods.

|  |  | $\chi_{para}$ | | | $|\chi_{dia}|$ | | |
|---|---|---|---|---|---|---|---|
|  |  | Frangi filter | GRE-based with $R_2^*$ | Proposed | Frangi filter | GRE-based with $R_2^*$ | Proposed |
| **DSC (%)** | $1 \times 1 \times 1$ mm³ (3T) | 55.8 ± 4.2 | 60.8 ± 1.8 | **76.7 ± 4.2** | 44.0 ± 5.9 | 53.2 ± 4.6 | **68.7 ± 7.9** |
|  | $0.65 \times 0.65 \times 0.65$ mm³ (7T) | 65.5 ± 3.4 | 66.4 ± 1.4 | **76.9 ± 2.7** | 55.6 ± 5.5 | 63.9 ± 3.8 | **72.6 ± 5.7** |

**Table 2**. DSCs for the four different $\chi$-separation algorithms.

|  |  | χ-sep-COSMOS | χ-sep-MEDI | χ-sep-iLSQR | χ-sepnet-$R_2^*$ |
|---|---|---|---|---|---|
| **DSC (%)** | $\chi_{para}$ | 76.7 ± 4.2 | 72.0 ± 3.8 | 72.0 ± 4.0 | 73.6 ± 4.1 |
|  | $|\chi_{dia}|$ | 68.7 ± 7.9 | 63.9 ± 7.4 | 63.7 ± 7.2 | 64.5 ± 7.1 |

**Table 3.** Reconstruction quality of $\chi$-sepnet-$R_2^*$ with respect to $\chi$-sep-COSMOS when analyzed with, without, and within the vessel mask.

|  | $\chi_{para}$ | | | $|\chi_{dia}|$ | | |
|---|---|---|---|---|---|---|
|  | RMSE (↓) | PSNR (↑) | SSIM (↑) | RMSE (↓) | PSNR (↑) | SSIM (↑) |
| **χ-sepnet-$R_2^*$ (without vessel mask)** | 0.0154 ± 0.0015 | 36.3 ± 0.8 | 0.926 ± 0.007 | 0.0145 ± 0.0012 | 36.8 ± 0.7 | 0.920 ± 0.009 |
| **χ-sepnet-$R_2^*$ (with vessel mask)** | **0.0121 ± 0.0011** | **38.4 ± 0.8** | **0.927 ± 0.007** | **0.0118 ± 0.0010** | **38.6 ± 0.8** | **0.922 ± 0.008** |
| **χ-sepnet-$R_2^*$ (within vessel mask)** | 0.0540 ± 0.0061 | 25.4 ± 1.0 | 0.894 ± 0.009 | 0.0512 ± 0.0060 | 25.9 ± 1.0 | 0.840 ± 0.016 |

**Table 4.** Application for the population-averaged ROI analysis. This table reports the proportion of vessels, and the mean susceptibility in each ROI analyzed with and without vessel masks.

| | | | Vessel portions [%] | Population average of mean [ppb] | | p-value |
|---|---|---|---|---|---|---|
| | | | | without vessel mask | with vessel mask | |
| $\chi_{para}$ | Subcortical nuclei | Caudate | 3.76 ± 1.67 | 47.5 ± 7.2 | 44.4 ± 6.8 | < 0.0001 * |
| | | Putamen | 0.17 ± 0.26 | 82.2 ± 21.5 | 82.1 ± 21.5 | < 0.0001 * |
| | | Globus pallidus | 0.02 ± 0.14 | 126.2 ± 16.1 | 126.1 ± 16.2 | 0.3276 |
| | | Nucleus accumbens | 0.01 ± 0.02 | 57.9 ± 15.0 | 57.9 ± 15.0 | 0.1539 |
| | | Substantia nigra | 0.01 ± 0.07 | 111.6 ± 17.2 | 111.5 ± 18.4 | 0.3261 |
| | | Red nucleus | 0.00 ± 0.00 | 104.4 ± 17.2 | 104.4 ± 17.2 | - |
| | | Ventral pallidum | 0.01 ± 0.07 | 134.9 ± 26.5 | 134.7 ± 27.0 | 0.2497 |
| | | Subthalamic nucleus | 0.00 ± 0.00 | 103.4 ± 15.2 | 103.4 ± 15.2 | 0.3197 |
| | Thalamic nuclei | Medial thalamic nuclei | 0.30 ± 0.19 | 34.8 ± 7.8 | 31.0 ± 7.4 | < 0.0001 * |
| | | Lateral thalamic nuclei | 0.03 ± 0.06 | 22.6 ± 5.1 | 22.3 ± 5.0 | < 0.0001 * |
| | | Pulvinar | 0.02 ± 0.14 | 50.8 ± 11.9 | 50.7 ± 12.0 | 0.1329 |
| $|\chi_{dia}|$ | White matter | Genu of corpus callosum | 1.89 ± 0.65 | 32.2 ± 3.0 | 30.7 ± 2.9 | < 0.0001 * |
| | | Body of corpus callosum | 0.66 ± 0.39 | 35.1 ± 2.6 | 34.7 ± 2.6 | < 0.0001 * |
| | | Splenium of corpus callosum | 0.53 ± 0.36 | 42.4 ± 4.0 | 42.2 ± 4.0 | < 0.0001 * |
| | | Cerebral peduncle | 0.16 ± 0.25 | 44.5 ± 3.9 | 44.2 ± 4.0 | 0.0002 * |
| | | Anterior limb of internal capsule | 0.13 ± 0.14 | 40.5 ± 4.1 | 40.3 ± 4.0 | < 0.0001 * |
| | | Posterior limb of internal capsule | 0.03 ± 0.06 | 52.1 ± 3.4 | 52.1 ± 3.4 | 0.0096 |
| | | Retrolentocular part of internal capsule | 0.09 ± 0.14 | 39.7 ± 3.5 | 39.6 ± 3.4 | < 0.0001 * |
| | | Anterior corona radiata | 0.02 ± 0.05 | 27.5 ± 2.8 | 27.5 ± 2.8 | 0.0016 * |
| | | Superior corona radiata | 0.03 ± 0.06 | 31.4 ± 2.9 | 31.4 ± 2.9 | < 0.0001 * |
| | | Posterior corona radiata | 0.02 ± 0.07 | 31.9 ± 2.6 | 31.8 ± 2.6 | 0.0719 |
| | | Posterior thalamic radiation | 0.04 ± 0.06 | 40.8 ± 4.4 | 40.8 ± 4.4 | < 0.0001 * |
| | | Sagittal stratum | 0.01 ± 0.02 | 36.8 ± 3.9 | 36.8 ± 3.9 | 0.0005 * |
| | | Superior longitudinal fasciculus | 0.01 ± 0.04 | 32.2 ± 2.8 | 32.2 ± 2.8 | 0.0650 |
| | Thalamic nuclei | Medial thalamic nuclei | 0.24 ± 0.20 | 14.5 ± 5.9 | 13.0 ± 5.3 | < 0.0001 * |
| | | Lateral thalamic nuclei | 0.03 ± 0.06 | 24.0 ± 4.7 | 23.9 ± 4.7 | < 0.0001 * |
| | | Pulvinar | 0.01 ± 0.09 | 6.3 ± 4.3 | 6.2 ± 4.2 | 0.1217 |

* p < 0.05 with Bonferroni-correction